\pdfoutput=1
\documentclass[11pt]{article}
\usepackage[review]{EMNLP2023}
\usepackage[export]{adjustbox}
\usepackage{enumitem}
\usepackage{booktabs} 
\usepackage{pgfplots}
\usetikzlibrary{pgfplots.groupplots}
\pgfplotsset{compat=1.3}
\usepackage{tikz}
\usetikzlibrary{shapes,arrows}
\usepackage{hyperref}

\usepackage{tabularx}

\usepackage{times}
\usepackage{latexsym}
\usepackage{graphicx}
\usepackage[T1]{fontenc}

\usepackage[utf8]{inputenc}
\usepackage{floatrow}
\usepackage{pifont}
\usepackage{microtype}

\usepackage{inconsolata}
\usepackage{cuted}
\usepackage{algorithm}
\usepackage{algpseudocode}

\title{Mixture of Rationale: Multi-Modal Reasoning Mixture for Visual Question Answering}

\author{
  Tao Li \\
  \texttt{taoli1@microsoft.com} \\\And
  Linjun Shou \\
  \texttt{lisho@microsoft.com} \\\And
  Xuejun Liu \\
  \texttt{xuejun.liu@nuaa.edu.cn}
}

\begin{document}
\nolinenumbers
\maketitle

\begin{abstract}
 Zero-shot visual question answering (VQA) is a challenging task that requires reasoning across modalities. While some existing methods rely on a single rationale within the Chain of Thoughts (CoT) framework, they may fall short of capturing the complexity of the VQA problem. On the other hand, some other methods that use multiple rationales may still suffer from low diversity, poor modality alignment, and inefficient retrieval and fusion. In response to these challenges, we propose \emph{Mixture of Rationales (MoR)}, a novel multi-modal reasoning method that mixes multiple rationales for VQA. MoR uses a single frozen Vision-and-Language Pre-trained Models (VLPM) model to {dynamically generate, retrieve and fuse multi-modal thoughts}. We evaluate MoR on two challenging VQA datasets, i.e. NLVR2 and OKVQA, with two representative backbones OFA and VL-T5. MoR achieves a 12.43\% accuracy improvement on NLVR2, and a 2.45\% accuracy improvement on OKVQA-S( the science and technology category of OKVQA). We make our code publicly available. \footnote{\url{https://www.bing.com}}
\end{abstract}

\section{Introduction}

\begin{figure}[t]
  \begin{center}
   \includegraphics[width=1\columnwidth]{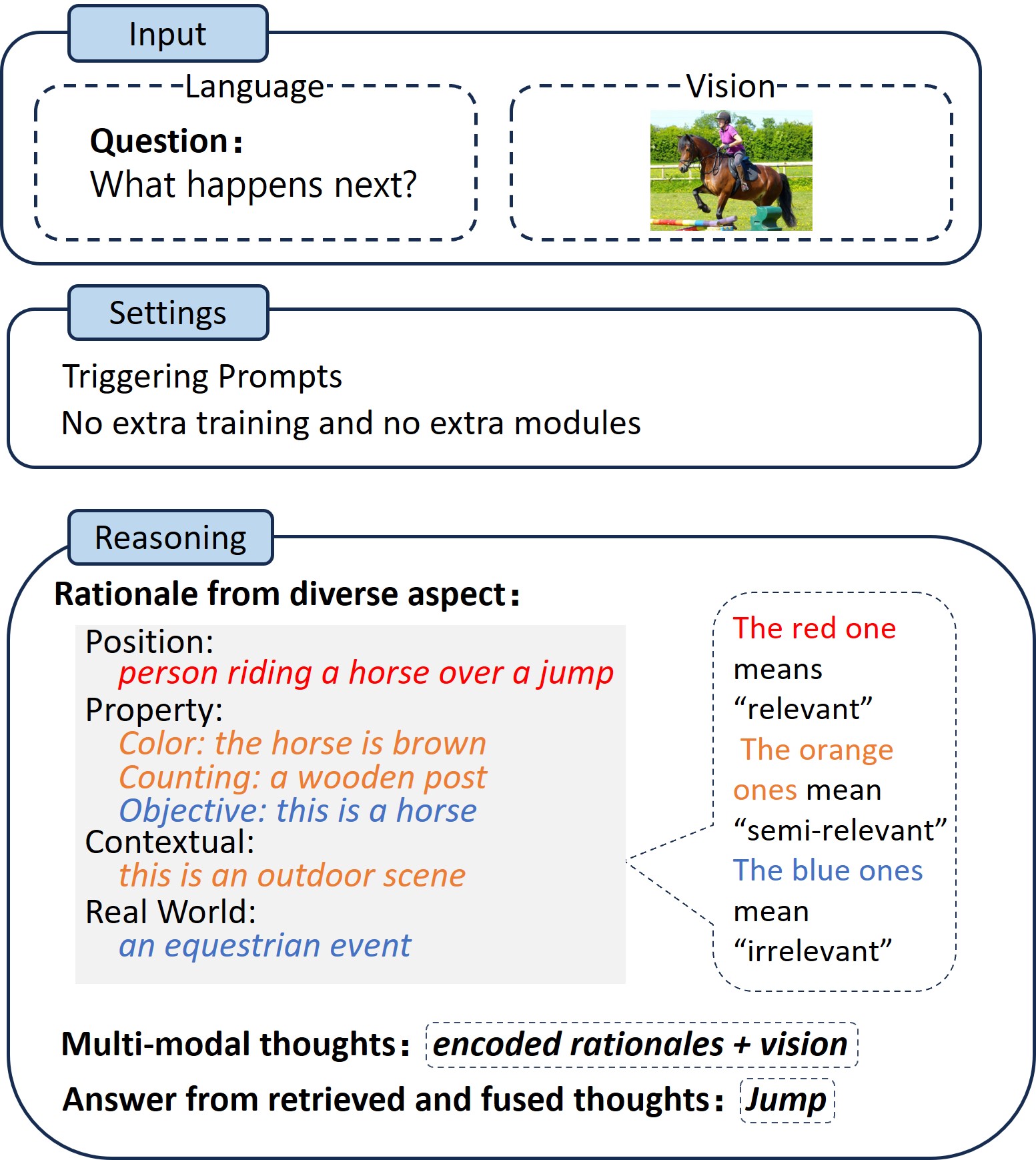}
  \end{center}
  \vspace{-0.6mm}
  \caption{A typical zero-shot visual question answering problem requires generating, retrieving and fusing multi-modal thoughts.}
    \vspace{-2mm}
  \label{fig_examples}
\end{figure}
Addressing the complexity of Zero-shot VQA \cite{Teney_Hengel_2016} presents a formidable challenge, demanding sophisticated multi-modal reasoning to bridge the gap between modalities lacking specific training. A concrete example of this intricate process is depicted in Figure \ref{fig_examples}, which involves a series of crucial steps: (1) Generating diverse rationales, (2) Producing multi-modal representations through the encoding of both rationales and visual stimuli, and (3) Discerning and filtering out irrelevant information, such as the equestrian events highlighted in the figure, and subsequently fusing the retrieved thoughts, depicted through the distinct colors of red and orange.

Many existing methods \cite{mmcot,yao2023beyond} rely on the CoT framework \cite{chu2023survey}, which uses VLPMs \cite{survey-vlp} as backbones and generates one single rationale for each answer. However, while CoT offers a solid foundation\cite{mmcot}, it may not adequately address the intricacies of the VQA challenge with such one single rationale, potentially resulting in suboptimal rationales or even responses that stray from the intended query.

Some other efforts have tried to improve the CoT framework on LLMs (Large Language Models), such as Tree of Thoughts (ToT) \cite{tot} and Graph of Thoughts (GoT) \cite{got}, which offer dynamic acquisition and aggregation of intermediate thoughts pertinent to the problem during inference. However, a significant drawback is their confinement to a single modality. Moreover, the generation of diverse rationales may be hindered by dependence on either domain knowledge for proposing thoughts or a singular prompt for sampling thoughts.

\if
In multi-modal areas, there are also some methods that use captions as rationales to prompt LLMs or LMs (Language Models). These methods were first introduced by PICa ~\cite{pica} and have two aspects of improvement. One is question-rationale relevance \cite{hu2022promptcap}, which evaluates how well the rationales are related to the problem. Another one is answer groundedness \cite{img2prompts}, which measures how well they support the answer.

However,
these methods have some limitations: first, they use a single prompt to generate all the rationales, which reduces the diversity and quality of the rationales \cite{naik2023diversity}. Second, they produce single-modal thoughts that still hinder the alignment of modalities. Third, they have difficulties in retrieving or fusing the relevant thoughts.
\fi

In this paper, we focus on the task of Zero-shot VQA, where intricate reasoning is essential to bridge the gap between modalities lacking specific training. Drawing inspiration from the CoT framework \cite{chu2023survey}, we introduce the Mixture of Rationales (MoR) as a novel approach to multi-modal reasoning in zero-shot VQA, which aims to tackle the aforementioned limitations. In particular, MoR leverages a frozen pre-trained Vision-and-Language Pre-trained Model (VLPM) to dynamically generate, retrieve, encodes and fuse thoughts, thereby facilitating comprehensive reasoning across modalities. In general, our main contributions are as follows. 
\begin{itemize}
\item  We introduce MoR, a novel model designed to tackle multi-modal reasoning for zero-shot VQA, which requires only a single frozen VLPM.

\item  Our MoR leverages the mixture of rationales, which provides diverse rationales and multi-modal thoughts. By dynamically retrieving and integrating these thoughts, a more precise answer can be obtained.

\item We evaluate MoR on two benchmark datasets, NLVR2 and OKVQA-S, demonstrating its effectiveness. In NLVR2 \cite{nlvr2}, we achieve a notable accuracy improvement of 12.43\%, while in OKVQA-S \cite{okvqa}, the improvement is 2.45\%.

\end{itemize}

\begin{table*}[htb]
    \centering\small
   \caption{Comparative summary of VQA approaches in different segments: (1) VLPMs, (2) fine-tuning methods, (3) caption-based methods, and (4) our proposal. The 'Rationales' column is classified by quantity and relevance to the problem (\ding{73} single/unrelated, \ding{75} single/related, \ding{72} multiple/related). 'Models' abbreviates key processes (RG=Rationale Generation, AG=Answer Generation, IQM=Image-Question Matching, C=Captioning, QG=Question Generation; FT=fine-tuning required).}

    \label{tab:mm+methods}
    \setlength{\tabcolsep}{6pt}
    {
        \begin{tabular}{lllllllll}\toprule
            {Models} & {Thoughts} & {Retrieval}  & {Fusion}  & {Models} & {Inference} & {Stages} & {FT-models}\\
            \midrule
            VLPMs  & \ding{73}  &  \ding{55} &  \ding{55} & VLPM & FT & 1 & 1\\
            \midrule
            MM-CoT  & \ding{73}   &  \ding{55} &  \ding{55} & FT-RG,  FT-AG& FT  & 2& 2\\
            MM-GoT & \ding{73}   &  \ding{55} &  \ding{55} & FT-RG, FT-AG& FT  & 2& 2\\
            \midrule
            PICa & \ding{73}  &  \ding{55} &  \ding{55} & C,LLM &Few-Shot & 2& 0\\
            PromptCap  & \ding{75} &  \ding{51}  &  \ding{55} & FT-C, LLM & Few-Shot  & 2& 1\\
            InterVQA & \ding{72} & \ding{51} &    MV & FT-C, Entailment, FT-RG & FT  & 3 & 2\\
            PNP-VQA  & \ding{72} & \ding{51}   &  FiD & C, FT-QA & FT  & 2 & 1\\
            Img2LLM  & \ding{72} & \ding{51} &    LLM & IQM, C, AG, FT-QG, LLM & Few-Shot  & 5 & 1\\
            \midrule
            Ours  & \ding{72} & \ding{51} &  \ding{51} & VLPM & Zero-Shot  & 2 & 0\\
            
            \bottomrule
        \end{tabular}
    }
\end{table*}

\section{Related Work}
\subsection{VLPMs}
Vision-and-Language Pre-trained Models (VLPMs) \cite{survey-vlp} are powerful backbone models that can handle various tasks across different modalities. They have developed into many unified generative models since the emergence of transformer \cite{vaswani2017attention} and CLIP \cite{radford2021learning}. These models can have different architectures, such as (1) encoder-only \cite{beitv1, beitv2, beitv3}, (2) or encoder-decoder \cite{ofa, vl-t5}, (3) or a mixture of encoder-decoder \cite{blipv1, blipv2}.

VLPMs provide a solid basis for exploring multi-modal reasoning. Researchers can exploit their general ability. For example, recent works of \cite{mmcot,yao2023beyond} have fine-tuned models to generate rationales and answers, implementing CoT. Nevertheless, the practice of generating a single rationale for each problem is increasingly recognized as insufficient for the VQA problem.

\subsection{XoT}
XoT (CoT, ToT, and GoT) \cite{chu2023survey} is a family of prompting techniques that use intermediate rationales to guide LLMs to solve complex NLP tasks. CoT was the first technique proposed by \cite{wei2022chain}, which showed its effectiveness on various reasoning tasks. ToT improved CoT by allowing LLMs to explore and backtrack multiple thought branches for each problem, creating a tree structure. GoT further generalized CoT and ToT by representing the generated thoughts as a graph. These techniques improved the reasoning abilities of LLMs significantly, as they resemble human thinking mechanisms.

However, they also have some limitations, such as the diversity of the generated thoughts. Although ToT and GoT mitigated the limitation of diversity by sampling or proposing rationales, these methods either require domain expertise when proposing or rely on a single source of information when sampling. According to \cite{naik2023diversity}, using different sources can increase the diversity of thoughts, which XoT may lack. Moreover, these techniques are only applicable to a single modality.

\subsection{Zero-Shot VQA}
VLPMs and CoTs bring zero-shot VQA new possibilities.  Several recent works have shown the potential of zero-shot VQA using caption-based methods. PICa \cite{pica} demonstrated the feasibility of this approach, but its rationales suffered from misalignment problems.

Two notable methods, PromptCap \cite{hu2022promptcap} and PNP-VQA, improved the alignment by question-rationale relevance. PromptCap fine-tunes a question-aware caption model, yet it provides a single rationale per problem, which may not fully capture the complexity of VQA tasks. PNP-VQA \cite{pnp} extract question-related patches, generating captions for these and then fusing them with Fusion in Decoder (FiD) only in language level.

Extending PNP-VQA, Img2LLM \cite{img2prompts} further improved answer groundedness. Following the generation of captions, it proceeds to produce QA pairs as exemplars for the In-Context-Learning \cite{dong2022survey}. The caption-only rationales are not sufficiently diverse to improve reasoning, and these methods requires 3 to 6 addtional models according to table \ref{tab:mm+methods}.

Another innovative  approach, Intervqa\cite{fu2023interpretable}, applied reverse thinking to retrieve and fuse relevant rationales. But the rationales generator needs to be fine-tuned and answers are not open-ended. It relies on predefined answers to get gold rationales, and then constructs reasoning graphs from generated clues to these rationales, followed by a majority vote to determine the final answer.

In summary, existing methods lack some important functionalities, such as {(1) diverse rationales, (2) retrieval or fusion multi-modal thoughts}, and may need extra fine-tuned models. In this paper, we propose an effective system for zero-shot VQA that overcomes these limitations.

\section{Preliminary}
\subsection{Notations}
We {formalize} VQA problem and our methods that employ encoder-decoder VLPM for problem-solving. Consider a set of \( N \) problems, for which the inputs are specified as follows:

\begin{itemize}
\item Let $\mathbf{Q} = \{q_1, \cdots, q_N\} $ represent the questions and $\mathbf{V} = \{V_1, \cdots, V_N\}$ denote the corresponding images for the problems.
\item $\mathbf{T} = \{T_1, \cdots, T_n\}$: A fixed set of triggering prompts.
\end{itemize}

The process involves the use of intermediate variables:

\begin{itemize}
\item $\mathbf{R} = \{R_1, \cdots, R_N\}$: A set of rationales, which derive $\mathbf{R'}$ by a set of link word $\mathbf{T'}$.
\item Eeach \( Z_i \), the embeddings \( z_1, \ldots, z_n \) are generated by ${R'_i}$ via the encoder $Enc$. The embedding $z_0$ is the initial thought produced without rationales.
\item $\mathbf{Z} = \{Z_1, \cdots, Z_N\}$: Multi-modal thoughts. These are encoded as embeddings.
\item $\mathbf{S} = \{S_1, \cdots, S_N\}$: Similarities between ${Z_i}$ and $z_0$.
\end{itemize}

The final output  \( y \) is generated by decoder $Dec$ from a subset of $\mathbf{Z}.$

\subsection{Fusion in Decoder}
\label{sec:fid}

 The Fusion in Decoder (FiD) algorithm is originally a technique \cite{izacard-grave-2021-leveraging} in T5 \cite{raffel2020exploring}  using retrieved passages to boost the performance of language models on various natural language processing tasks. It uses T5 encoder-decoder architecture, which is not the only choice, where the encoder produces embeddings from a question and multiple passages, and then the decoder generates an answer based on the fused embeddings. 
 
 In our paper, we have extended this approach to multi-modalities: based on VLPM's encoder and decoder, concatenating retrieved embeddings  $z_1, \cdots, z_k$ before the decoder generates the final answer.

\begin{figure*}[!ht]
\centering
\includegraphics[width=1.0\textwidth, center]{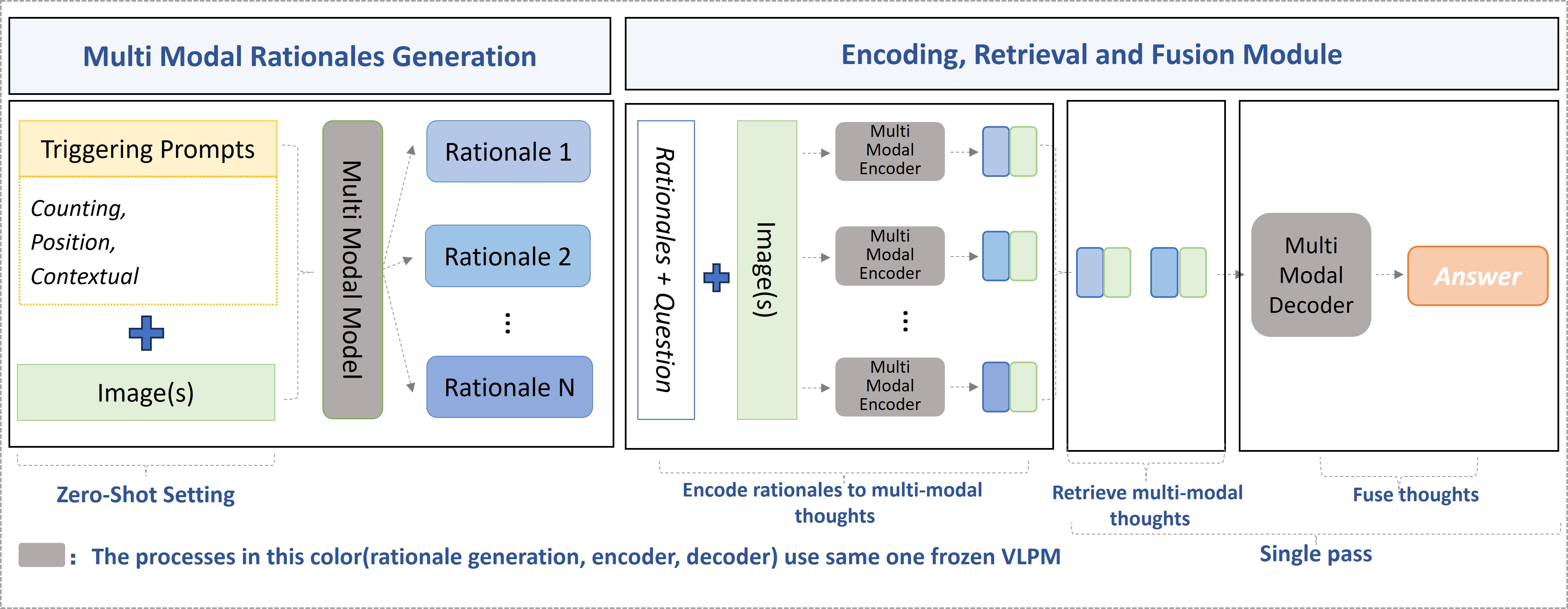}
\setlength{\abovecaptionskip}{-10pt}
\setlength{\belowcaptionskip}{-1000pt}
\caption{Diagram of MoR, which is based on any frozen encoder-decoder VLPM. The first module produces rationales from the input of triggering prompts, questions, and image(s). The second module performs encoding, retrieval and fusion in one pass.}
\label{fig:system}
\end{figure*}

\section{Methodology}

An overview of our work can be seen in Figure \ref{fig:system}. We will introduce our system in two parts. The first section involves the \hyperref[sec:reg]{\textcolor{blue}{{rationale generation}}}, which generate diverse rationales. The second section is about \hyperref[sec:rfm]{\textcolor{blue}{encoding, retrieving and fusing}} to get the final answer.




\subsection{Rationales Generation Module}
\label{sec:reg}
To enhance the VQA's rationales generation, we analyzed the \emph{Rationales} column in Table \ref{tab:mm+methods}.  This analysis highlighted two key requirements: (1) rationales should be relevant to the problem, (2)  rationales should come from different sources to enhance diversity, a concetp supported by the idea of \cite{naik2023diversity}. To this end, we initiate the rationales generation with triggering prompts $\mathbf{T}$ that can cover different aspects of the problem, having three steps: 
\begin{enumerate}

 \item Generating generic rationales $\{R_1, \cdots, R_m\}$: Similar to those utilized in \cite{wei2022chain},  these triggering prompts include generic ones like "\emph{Let’s think step by step}"
 \item Deriving specific rationales $\{R_m+1, \cdots, R_n\}$: These prompts take into account the contextual information related to the problem using prompts such as \textit{"What’s the scenario?"}.
 
 They also consider various properties like position, count, and color. For instance, triggering prompts can be structured as  "\emph{Let’s consider \#<object>}", where we can get \emph{\#<object>} by off the shelve key phrase extraction (KPE) model.

\item Formulation of intermediate rationales $\mathbf{R'}$.

Every rational $R_i$ is expanded into intermediate rationales $R_i'$ by concatenating triggering prompts $T_i$ and connecting word $T'$, forming $\{ T, R, T',q\}$. The connecting word $T'$ includes '\emph{Consequently}', '\emph{Therefore}','\emph{Then}', and so on.

\end{enumerate}

By using triggering prompts $\mathbf{T}$ we can generate rationales that are not only potentially related to the problem but also more diverse. 

    


\subsection{Encoding, Retrieval and Fusion Module}
\label{sec:rfm}

{Retrieval} and {Fusion} modules have potential to better solve VQA problems according to third segment of Table \ref{tab:mm+methods}, With this insights, our goal is to use intermediate rationales $\mathbf{R'}$ that produced in the preceding stage to dynamically encode, retrieve and fuse them. We achieve this in four steps:

\begin{enumerate}
    
    \item Generate multi-modal embeddings \(\mathbf{Z}\) by processing each intermediate prompt  $\{R_1, \cdots, R_n\}$ along with the images through the encoder \(Enc\).
    
    These embeddings represent the multi-modal thoughts. $z_i^0$ is generated solely from the query and image(s), serving as the representation of the problem itself.
    \item Compute cosine similarities between each thought $\mathbf{Z_i}$ and $\mathbf{z_i^0}$ to obtain a similarity score $\mathbf{S}$.
    \item Refine the thought set to the $k$ most relevant elements, $\mathbf{Z'_i}$, based on the similarity scores $\mathbf{S}$.
    \item Integrate $\mathbf{Z'_i}$ into a singular fused thought $z'_i$ employing the Fusion-in-Decoder (\hyperref[sec:fid]{\textcolor{blue}{FiD}}) approach. Alternatively, Majority Voting can be utilized as a fusion technique.
    \item Use $z'_i$ to generate the answer $y$ via $Dec$.
\end{enumerate}

Additionally, as seen in the column of \emph{Models} in table \ref{tab:mm+methods}  we use one backbone for the system. This implies that in this stage we use the same encoder-decoder VLPM backbone from the rationales generation. This aligns the multi-modal thoughts $\mathbf{Z}$ within the same representational space as the problem and answer. We have verified this alignment in the ablation study of the \hyperref[sec:exp_encoding_effect]{\textcolor{blue}{retrieval's model}}.


\section{Experiments}
In this section, we will report the significant gains over different combinations of backbones and tasks using our methods. Additionally, we will delve into detailed analyses.

\subsection{Dataset} 
We chose NLVR2\cite{nlvr2} and OKVQA-S\cite{okvqa} ( the science and technology category of OKVQA) as our datasets because they both pose different challenges for multi-modal reasoning. NLVR2 requires the model to compare, locate, and count objects based on the given question and images. OKVQA challenges the model to use external knowledge, such as common sense, and facts, to answer questions about different topics and scenarios.

NLVR2 is easy to fine-tune and achieve high performance, but it is very hard to perform well in the zero-shot setting. It is worth noting that the model may perform worse than random guessing on NLVR2, as it can generate answers other than ‘yes’ or ‘no’.

For OKVQA, We focus on OKVQA-S because it is a challenging category, and we will show that more sophisticated reasoning is especially helpful for this category to achieve higher performance.

\subsection{Baselines} 
\label{sec:baseline}
We can compare our approach with these baselines:
\begin{itemize} 
    \item The SOTA on NLVR2 is high at 92.58 as reported by \cite{beitv3}. Yet, when it comes to zero-shot reasoning on NLVR2, the baseline accuracy only stands at {50.14}, achieved by VL-T5 vanilla.
    \item In OKVQA, the SOTA model is VL-T5 vanilla having a score of {33.33}. 
    \item The best score in OKVQA-S is {31.09} by {VL-T5 vanilla}.
\end{itemize}

In terms of the PNP-VQA score claimed to be SOTA in its respective study, there are several points to consider:

\begin{itemize}
    \item The highest reported score is 35.9, based on the 11B model version. We evaluate the PNP-VQA base version, which has 223M parameters, comparable in size to OFA vanilla and VL-T5 vanilla.
    \item When comparing models with similar parameter counts, VL-T5 vanilla and OFA vanilla outperform the PNP-VQA base model, due to their use of the VQA-V2 training set during pre-training.
\end{itemize}

\subsection{Implementation} 
Our methods are based on VLPMs backbones, which have the potential for zero-shot tasks that are often ignored by researchers who prefer to fine-tune them. OFA and VL-T5 are two representative ones that have encoder-decoder architecture, with model sizes of 182M and 220M parameters respectively. Their open-source framework allows us to implement our methods easily and efficiently. We can write rationale generation, encoding, retrieval and fusion within the OFA or VL-T5 source code.

\subsection{Experiment Results and Analysis}
\begin{table}[htb]
    \centering\small
      \vspace{-3.6mm}
        \caption{A summary of the main results. {-w/CoT} denotes that the system only uses highest-scoring generated rationales only. Segment 1 and 2 are the backbones of {OFA} and {VL-T5} respectively. Segment 3 shows SOTA models for each task, elaborated in the \hyperref[sec:baseline]{\textcolor{blue}{Baseline}}. The best performances are in \textbf{bold}.}
        \label{tab:pre_position}
         \setlength{\tabcolsep}{4pt}
         {
              \begin{tabular}{lllcc}\toprule
                 {Method} & {NLVR2} & {OKVQA-S} & { OKVQA} \\
                 \midrule
                 OFA-Vanilla & 47.37 & 28.69  & \textbf{31.98}\\
                 OFA-w/ CoT  & 53.18 & 29.50  & 29.00\\
                 OFA-w/ MoR  & \textbf{60.80} & \textbf{31.14} & 29.79 \\
                 \midrule
                 VL-T5-Vanilla & 50.14 & 31.09 & \textbf{33.33} \\
                 VL-T5-w/ CoT  & 51.18 & 32.65  & 31.10 \\
                 VL-T5-w/ MoR  & \textbf{51.31} & \textbf{33.51} & 33.26 \\
                 \midrule
                  SOTA   & 50.14  & 31.09 & 31.33 \\

                 \bottomrule
             \end{tabular}
        \vspace{-1.8mm}
        }
 \label{tab:main_results}
\end{table}

In this section, we present and analyze the results from our various experiments, including our \hyperref[sec:exp-main]{\textcolor{blue}main results}, an ablation study on \hyperref[sec:exp-re]{\textcolor{blue}rationale generation}, and experiments on \hyperref[sec:exp-retrie]{\textcolor{blue}encoding, retrieval and fusion module}. Additionally, in \hyperref[sec:casestudy]{\textcolor{blue}{Appendix A}}, we present a case study that shows how each component works together.

\subsubsection{Main Result}
\label{sec:exp-main}
Table \ref{tab:main_results} shows that MoR improves OFA and VL-T5 on NLVR2 and OKVQA-S significantly.
\begin{itemize}
    \item It increases OFA by {12.43\%} points on NLVR2 and {2.45\%} points on OKVQA-S, 
    \item It raises VL-T5 by {1.57\%} points on NLVR2 and {2.42\%} points on OKVQA-S. 

\end{itemize}

Vanillas do not benefit from MoR on OKVQA because some categories do not require complex reasoning. However, it is needed for OKVQA-S, which is the category of science and technology. We provide the category-wise results in the \hyperref[sec:ea-okvqa]{\textcolor{blue}{Appendix B}}.

The main results show that MoR is an effective technique to improve the reasoning abilities of VLPMs.

\subsubsection{Rationales Generation Module} 
\label{sec:exp-re}
The influence of the rationales generation is verified in this section.
\paragraph{Effect of Rationales}
The model labeled '-w/ CoT' in each segment of table \ref{tab:main_results} shows the results of using a single rationale to generate multi-modal thoughts outperforms the baseline models on both tasks. The improvements are as follows:
\begin{itemize}
    \item For NLVR2, it achieves an improvement of {5.81} points over OFA, and a improvement of {1.04} points over VL-T5.
     \item For OKVQA-S, it improves the accuracy by {0.91} points over OFA and {0.56} points over VL-T5.
\end{itemize}

\paragraph{Diversity of Rationales}

\begin{figure}[t]
  \begin{center}
   \includegraphics[width=1\columnwidth]{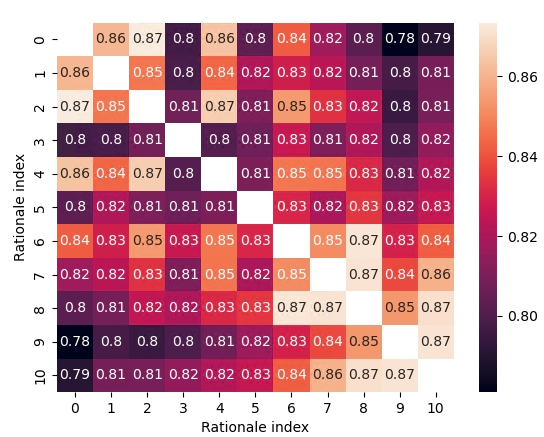}
  \end{center}
  \vspace{-0.6mm}
  \caption{Diversity of rationales for NLVR2 task. The figure shows the average cosine similarity between each pair of rationales. Similarities are computed using OpenAI’s text-embedding-ada-002 model \cite{openai2022textembeddingada}. It can be observed that there are two clusters that have high intra-similarity (lighter color) and low inter-similarity(darker color), indicating diversity. The index of the rationales can be found in \hyperref[sec:rationale_index]{\textcolor{blue}{Appendix C}}. }
    \vspace{-2mm}
  \label{fig_heatmap}
\end{figure}

Figure \ref{fig_heatmap} illustrates the degree of similarity among various rationales. The rationales are divided into two groups as described in the section on \hyperref[sec:rationale_index]{\textcolor{blue}{Appendix C}}. The first group, with labels from 6 to 10, comprises generic rationales. The second group, with labels from 0 to 5, consists of specific rationales.

\begin{figure}[t] 
\centering 
\pgfplotsset{height=5.2cm,width=7.68cm,every axis/.append style={thick},every tick label/.append style={font=\small},every axis legend/.append style={at={(0.5,0.95)}},legend columns=1} 
\begin{tikzpicture} 
\tikzset{every node}=[font=\small] 
\begin{axis}[
    align=center,
    legend style={at={(0.5,1.3)},anchor=north},
    ymin=0.75, ymax=0.9,
    xticklabels={0,1,2,3,4,5,6,7,8,9,10},
    xtick={0,1,2,3,4,5,6,7,8,9,10},
    ylabel style={align=center},
    xlabel={Rationale Index},
    ylabel={Cosine similarity},
    ytick={0.75, 0.77, 0.79, 0.81, 0.83, 0.85, 0.87, 0.89},
    ymajorgrids=true,
    xmajorgrids=true,
    grid style=dashed,
    xtick pos=bottom,
    ytick pos=left,
]

\addplot+ [color=blue, mark=o, mark size=2.5pt, ] coordinates {
(0, 0.8556) (1, 0.8289) (2, 0.8284) (3, 0.8208) (4, 0.8110) (5, 0.7999) (6, 0.7938) (7, 0.7898) (8, 0.7835) (9, 0.7833) (10, 0.7691)

};
\addlegendentry{Rationale-Problem Similarity}

\addplot+ [color=red, mark=triangle, mark size=2.5pt, ] coordinates {
    (1, 0.8707) (2, 0.8781) (3, 0.8797) (4, 0.8813) (5, 0.8840) (6, 0.8845) 
    (7, 0.8838) (8, 0.8836) (9, 0.8840) (10, 0.8822)
};
\addplot+[color=red, only marks, mark=triangle*, mark size=4pt] coordinates {
    (6, 0.8845)
};
\addlegendentry{Mean Pooling Rationale-Problem Similarity}

\end{axis} 
\end{tikzpicture} 
\caption{Similarities between problems and rationales. The blue line shows how similar each rationale is to the problem, as measured by cosine similarity using the text-embedding-ada-002 model. The red line indicates the similarity between the problem and the average of all previous rationales.  The rationales are numbered in \hyperref[sec:rationale_index]{\textcolor{blue}{Appendix C}}.}

\label{fig:comparison}
\end{figure}
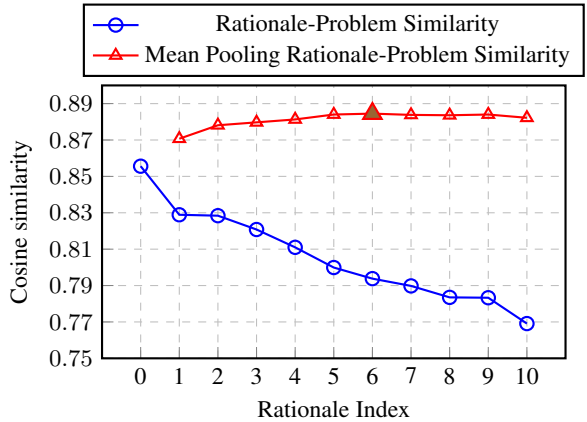

The color gradient conveys the average cosine similarity between pairs of rationales. We can observe that the rationales within each category are very similar to each other, but not to the other category. Additionally, figure \ref{fig:comparison} reveals how similarities between rationales and given problem changes with different rationales.

Therefore, we can infer triggering prompts can bring the diversity of the rationales.

\paragraph{Potential of Rationales}

An analysis of the similarity between rationales and problems illustrate the potential of the rationales, even when considered purely at the linguistic level.  As depicted in Figure \ref{fig:comparison}, the red line means the fusion rationale similarity to the question, while the blue line corresponds to no fusion. 

The red line consistently outperforms the blue line demonstrating the fusion's potential. Specifically, mean pooling the top six rationales yields the highest similarity score, indicating that retrieval will significantly benefits the system.

\subsubsection{Encoding, Retrieval and Fusion Module}
\label{sec:exp-retrie}

In this section,  we evaluate the effectiveness of the encoding, retrieval and fusion module.

\paragraph{Effect of Retrieval}
The effect of the retrieval component is assessed using the NLVR2 task, as shown in Table \ref{tab:reason_results}. The observations are as follows:
\begin{itemize}
    \item There is an increase in accuracy by {0.8} points when using the retrieval component, by comparing the last third row with the last second one in the table.
    \item  The last row shows that dynamic retrieval of rationales leads to {1.55} more improvement than using a fixed set of top 2 rationales.

\end{itemize}

\begin{table}[h] 
    \centering\small 
    \vspace{-3mm} 
    \caption{The effect of different modules on the score of NLVR2 when using OFA. MV Fusion means the Fusion method is Majority Voting. Dynamic Rationales indicate the rationales are dynamically selected during the inference. Without dynamic rationales implies the use of two predetermined best rationales.} 
    \label{tab:score_analysis} 
    \begin{tabular}{lcccc}
    \toprule 
    CoT & Retrieval & Fusion & Dynamic Rationales & Score \\
    \midrule 
    \ding{55} & \ding{55} & \ding{55} & \ding{55} & 47.37 \\
    \midrule 
    \ding{51} & \ding{55} & \ding{55} & \ding{55} & 53.18 \\    
    \ding{51} & \ding{55} & MV & \ding{55} & 55.16 \\
    \ding{51} & \ding{55} & FiD & \ding{55} & 58.45 \\
    \midrule 
    \ding{51} & \ding{51} & FiD & \ding{55} & 59.25 \\
    \ding{51} & \ding{51} & FiD & \ding{51} & \textbf{60.80} \\
    \bottomrule 
    \end{tabular} 
     \label{tab:reason_results}
\end{table}

\begin{figure}[t] 
\centering 
\pgfplotsset{height=5.2cm,width=7.68cm,compat=1.14,every axis/.append style={thick},every tick label/.append style={font=\small},every axis legend/.append style={ at={(0.5,0.95)}},legend columns=2 row=2} 
    \begin{tikzpicture} \tikzset{every node}=[font=\small] 
        \begin{axis} [ 
                align = center, 
                legend style={at={(0.5,1.3)},anchor=north}, 
                ymin=25, ymax=35, 
                xticklabels={1,2,3,4,5,6,7,8,9,10}, xtick={0,1,2,3,4,5,6,7,8,9}, 
                ylabel style={align=center}, xlabel={Topk}, ylabel={Score}, ytick={25, 27, 29, 31, 33, 35}, 
                ymajorgrids=true, xmajorgrids=true, grid style=dashed, xtick pos=bottom, ytick pos=left, 
            ]
        
        \addplot+ [color=blue, mark=o, mark size=2.5pt, ] coordinates {(0, 32.76) (1, 33.505) (2, 33.05) (3, 32.63) (4, 32.955) (5, 32.5) (6, 32.305) (7, 32.325) (8, 32.305)};
        \addlegendentry{Backbone: VL-T5}
        
       \addplot+ [color=blue, mark=triangle, mark size=2.5pt, ] coordinates {(0, 31.14) (1, 30.21) (2, 30.21) (3, 29.87) (4, 29.62) (5, 29.05) (6, 28.35) (7, 27.81) (8, 27.43)};
       \addlegendentry{Backbone: OFA}

        \end{axis} 
    \end{tikzpicture} \vspace{-1mm} 
\caption{The effect of varying the number of thoughts, that are dynamically retrieved, on our model’s performance for the OKVQA-S dataset.}
\label{fig:curve} \vspace{-2mm} 
\end{figure}
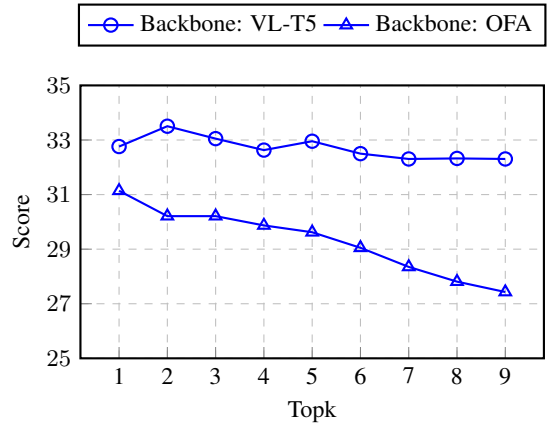

\paragraph{Model of Retrieval}
\label{sec:exp_encoding_effect}

\begin{table}[htb]
    \centering\small
      \vspace{-3.6mm}
        \caption{The performance of MoR on NLVR2 task using OFA with different retrieval methods. The methods are: (1) Mean, which uses mean pooling of token embedding to compute the similarity between rationales and problems; (2) CLS, which uses the classification token to compute the similarity; and (3) Contriever \cite{contriever}, which uses an external system to compute the similarity.}
         \setlength{\tabcolsep}{36pt}
         {
              \begin{tabular}{ll}\toprule
                 {Method} & {Score} \\
                 \midrule
                 Mean & 60.80 \\
                 CLS & 60.28 \\
                 Contrieval & 58.63 \\
                 \bottomrule
             \end{tabular}
        \vspace{-1.8mm}
        }
 \label{tab:retrieval_result}
\end{table}

 We compare different retrieval methods for our system on the NLVR2 dataset in Table \ref{tab:retrieval_result}. The best method is to use mean pooling of the encoder’s embeddings during the inference. 
 The reason is that the embeddings contain multi-modal information and consist of the answer. 
 
 On the other hand, external models, such as the unsupervised dense information retrieval model contriever \cite{contriever}, perform poorly,  due to their single modality and their inconsistency with the answer. This proves that encoding within the process of reasoning is crucial to the quality.

\paragraph{Number of Retrieved thoughts}

Figure \ref{fig:curve} illustrates how the score of OKVQA-S depends on the number of thoughts used for each problem. We dynamically select the best k for each problem. Different backbones have different optimal numbers. For instance, {k=1} is the best for OFA, while {k=2} is the ideal for VL-T5.

\paragraph{Effect of Fusion}

The fusion component integrates the thoughts that are retrieved dynamically. The data in the second part of Table \ref{tab:reason_results} illustrates the following improvements:

\begin{itemize}
    \item Implementing the fusion component with majority voting results in an accuracy boost of 1.98 points over the baseline.
    \item Employing FiD further improves the accuracy by an additional 3.29 points beyond what is achieved with majority voting.
\end{itemize}

\section*{Conclusion}

In this paper, we propose MoR, an effective system for multi-modal reasoning in zero-shot VQA. MoR has two modules based on only one VLPM backbone. The first one to generate diverse rationales from triggering prompts. The second module encodes, retrieves and fuses these thoughts to generate answers. We test MoR on two challenging datasets: NLVR2 and OKVQA-S, demonstrating that MoR significantly improves the zero-shot performance on both tasks with different backbone models such as OFA and VL-T5.

For future work, we plan to explore the following directions: (1) designing a system that can automatically generate triggering prompts and intermediate rationales; (2) applying backtracking and aggregating strategy in multi-modal reasoning such as the ToT \cite{tot} or GoT \cite{tot} to enhance the multi-modal's retrieval and fusion process; (3) experimenting with larger models such as GPT-V, Gemini \cite{gptv,team2023gemini} to further boost the reasoning performance.

\bibliography{anthology,custom}

\begin{thebibliography}{32}
\expandafter\ifx\csname natexlab\endcsname\relax\def\natexlab#1{#1}\fi

\bibitem[{Bao et~al.(2022)Bao, Dong, and Wei}]{beitv1}
H~Bao, L~Dong, and F~Wei. 2022.
\newblock Beit: Bert pre-training of image transformers. arxiv 2021.
\newblock \emph{arXiv preprint arXiv:2106.08254}.

\bibitem[{Besta et~al.(2023)Besta, Blach, Kubicek, Gerstenberger, Gianinazzi, Gajda, Lehmann, Podstawski, Niewiadomski, Nyczyk, and Hoefler}]{got}
Maciej Besta, Nils Blach, Ales Kubicek, Robert Gerstenberger, Lukas Gianinazzi, Joanna Gajda, Tomasz Lehmann, Michal Podstawski, Hubert Niewiadomski, Piotr Nyczyk, and Torsten Hoefler. 2023.
\newblock Graph of thoughts: Solving elaborate problems with large language models.
\newblock \emph{arXiv preprint arXiv:2308.09687}.

\bibitem[{Cho et~al.(2021)Cho, Lei, Tan, and Bansal}]{vl-t5}
Jaemin Cho, Jie Lei, Hao Tan, and Mohit Bansal. 2021.
\newblock Unifying vision-and-language tasks via text generation.
\newblock In \emph{International Conference on Machine Learning}, pages 1931--1942. PMLR.

\bibitem[{Chu et~al.(2023)Chu, Chen, Chen, Yu, He, Wang, Peng, Liu, Qin, and Liu}]{chu2023survey}
Zheng Chu, Jingchang Chen, Qianglong Chen, Weijiang Yu, Tao He, Haotian Wang, Weihua Peng, Ming Liu, Bing Qin, and Ting Liu. 2023.
\newblock A survey of chain of thought reasoning: Advances, frontiers and future.
\newblock \emph{arXiv preprint arXiv:2309.15402}.

\bibitem[{Dong et~al.(2022)Dong, Li, Dai, Zheng, Wu, Chang, Sun, Xu, and Sui}]{dong2022survey}
Qingxiu Dong, Lei Li, Damai Dai, Ce~Zheng, Zhiyong Wu, Baobao Chang, Xu~Sun, Jingjing Xu, and Zhifang Sui. 2022.
\newblock A survey for in-context learning.
\newblock \emph{arXiv preprint arXiv:2301.00234}.

\bibitem[{Fu et~al.(2023)Fu, Zhou, Chen, Yatskar, and Roth}]{fu2023interpretable}
Xingyu Fu, Ben Zhou, Sihao Chen, Mark Yatskar, and Dan Roth. 2023.
\newblock Interpretable by design visual question answering.
\newblock \emph{arXiv preprint arXiv:2305.14882}.

\bibitem[{Gan et~al.(2022)Gan, Li, Li, Wang, Liu, Gao et~al.}]{survey-vlp}
Zhe Gan, Linjie Li, Chunyuan Li, Lijuan Wang, Zicheng Liu, Jianfeng Gao, et~al. 2022.
\newblock Vision-language pre-training: Basics, recent advances, and future trends.
\newblock \emph{Foundations and Trends{\textregistered} in Computer Graphics and Vision}, 14(3--4):163--352.

\bibitem[{Greene et~al.(2022)Greene, Sanders, Weng, and Neelakantan}]{openai2022textembeddingada}
Ryan Greene, Ted Sanders, Lilian Weng, and Arvind Neelakantan. 2022.
\newblock New and improved embedding model.
\newblock OpenAI Blog.
\newblock Available: \url{https://openai.com/blog/new-and-improved-embedding-model/}.

\bibitem[{Guo et~al.(2023)Guo, Li, Li, Huat~Tiong, Li, Tao, and Hoi}]{img2prompts}
Jiaxian Guo, Junnan Li, Dongxu Li, Anthony~Meng Huat~Tiong, Boyang Li, Dacheng Tao, and Steven Hoi. 2023.
\newblock From images to textual prompts: Zero-shot visual question answering with frozen large language models.
\newblock In \emph{2023 IEEE/CVF Conference on Computer Vision and Pattern Recognition (CVPR)}, pages 10867--10877.

\bibitem[{Hu et~al.(2022)Hu, Hua, Yang, Shi, Smith, and Luo}]{hu2022promptcap}
Yushi Hu, Hang Hua, Zhengyuan Yang, Weijia Shi, Noah~A Smith, and Jiebo Luo. 2022.
\newblock Promptcap: Prompt-guided task-aware image captioning.
\newblock \emph{arXiv preprint arXiv:2211.09699}.

\bibitem[{Izacard et~al.(2021)Izacard, Caron, Hosseini, Riedel, Bojanowski, Joulin, and Grave}]{contriever}
Gautier Izacard, Mathilde Caron, Lucas Hosseini, Sebastian Riedel, Piotr Bojanowski, Armand Joulin, and Edouard Grave. 2021.
\newblock Unsupervised dense information retrieval with contrastive learning.
\newblock \emph{arXiv preprint arXiv:2112.09118}.

\bibitem[{Izacard and Grave(2021)}]{izacard-grave-2021-leveraging}
Gautier Izacard and Edouard Grave. 2021.
\newblock Leveraging passage retrieval with generative models for open domain question answering.
\newblock In \emph{Proceedings of the 16th Conference of the European Chapter of the Association for Computational Linguistics: Main Volume}, pages 874--880, Online. Association for Computational Linguistics.

\bibitem[{Li et~al.(2023)Li, Li, Savarese, and Hoi}]{blipv2}
Junnan Li, Dongxu Li, Silvio Savarese, and Steven Hoi. 2023.
\newblock Blip-2: Bootstrapping language-image pre-training with frozen image encoders and large language models.
\newblock \emph{arXiv preprint arXiv:2301.12597}.

\bibitem[{Li et~al.(2022)Li, Li, Xiong, and Hoi}]{blipv1}
Junnan Li, Dongxu Li, Caiming Xiong, and Steven Hoi. 2022.
\newblock Blip: Bootstrapping language-image pre-training for unified vision-language understanding and generation.
\newblock In \emph{International Conference on Machine Learning}, pages 12888--12900. PMLR.

\bibitem[{Marino et~al.(2019)Marino, Rastegari, Farhadi, and Mottaghi}]{okvqa}
Kenneth Marino, Mohammad Rastegari, Ali Farhadi, and Roozbeh Mottaghi. 2019.
\newblock Ok-vqa: A visual question answering benchmark requiring external knowledge.
\newblock In \emph{Proceedings of the IEEE/CVF Conference on Computer Vision and Pattern Recognition}, page 3195–3204.

\bibitem[{Naik et~al.(2023)Naik, Chandrasekaran, Yuksekgonul, Palangi, and Nushi}]{naik2023diversity}
Ranjita Naik, Varun Chandrasekaran, Mert Yuksekgonul, Hamid Palangi, and Besmira Nushi. 2023.
\newblock Diversity of thought improves reasoning abilities of large language models.
\newblock \emph{arXiv preprint arXiv:2310.07088}.

\bibitem[{OpenAI(2023)}]{gptv}
OpenAI. 2023.
\newblock Gpt-4v(ision) system card.
\newblock \emph{arXiv preprint arXiv:2309.17421}.

\bibitem[{Peng et~al.(2022)Peng, Dong, Bao, Ye, and Wei}]{beitv2}
Zhiliang Peng, Li~Dong, Hangbo Bao, Qixiang Ye, and Furu Wei. 2022.
\newblock Beit v2: Masked image modeling with vector-quantized visual tokenizers.
\newblock \emph{arXiv preprint arXiv:2208.06366}.

\bibitem[{Radford et~al.(2021)Radford, Kim, Hallacy, Ramesh, Goh, Agarwal, Sastry, Askell, Mishkin, Clark et~al.}]{radford2021learning}
Alec Radford, Jong~Wook Kim, Chris Hallacy, Aditya Ramesh, Gabriel Goh, Sandhini Agarwal, Girish Sastry, Amanda Askell, Pamela Mishkin, Jack Clark, et~al. 2021.
\newblock Learning transferable visual models from natural language supervision.
\newblock In \emph{International conference on machine learning}, pages 8748--8763. PMLR.

\bibitem[{Raffel et~al.(2020)Raffel, Shazeer, Roberts, Lee, Narang, Matena, Zhou, Li, and Liu}]{raffel2020exploring}
Colin Raffel, Noam Shazeer, Adam Roberts, Katherine Lee, Sharan Narang, Michael Matena, Yanqi Zhou, Wei Li, and Peter~J Liu. 2020.
\newblock Exploring the limits of transfer learning with a unified text-to-text transformer.
\newblock \emph{The Journal of Machine Learning Research}, 21(1):5485--5551.

\bibitem[{Suhr et~al.(2019)Suhr, Zhou, Zhang, Zhang, Bai, and Artzi}]{nlvr2}
Alane Suhr, Stephanie Zhou, Ally Zhang, Iris Zhang, Huajun Bai, and Yoav Artzi. 2019.
\newblock \href {https://doi.org/10.18653/v1/P19-1644} {A corpus for reasoning about natural language grounded in photographs}.
\newblock In \emph{Proceedings of the 57th Annual Meeting of the Association for Computational Linguistics}, pages 6418--6428, Florence, Italy. Association for Computational Linguistics.

\bibitem[{Team et~al.(2023)Team, Anil, Borgeaud, Wu, Alayrac, Yu, Soricut, Schalkwyk, Dai, Hauth et~al.}]{team2023gemini}
Gemini Team, Rohan Anil, Sebastian Borgeaud, Yonghui Wu, Jean-Baptiste Alayrac, Jiahui Yu, Radu Soricut, Johan Schalkwyk, Andrew~M Dai, Anja Hauth, et~al. 2023.
\newblock Gemini: a family of highly capable multimodal models.
\newblock \emph{arXiv preprint arXiv:2312.11805}.

\bibitem[{Teney and Hengel(2016)}]{Teney_Hengel_2016}
Damien Teney and Antonvanden Hengel. 2016.
\newblock Zero-shot visual question answering.
\newblock \emph{Cornell University - arXiv,Cornell University - arXiv}.

\bibitem[{Tiong et~al.(2022)Tiong, Li, Li, Savarese, and Hoi}]{pnp}
Anthony Meng~Huat Tiong, Junnan Li, Boyang Li, Silvio Savarese, and Steven~C.H. Hoi. 2022.
\newblock Plug-and-play {VQA}: Zero-shot {VQA} by conjoining large pretrained models with zero training.
\newblock In \emph{Findings of the Association for Computational Linguistics: EMNLP 2022}, pages 951--967, Abu Dhabi, United Arab Emirates. Association for Computational Linguistics.

\bibitem[{Vaswani et~al.(2017)Vaswani, Shazeer, Parmar, Uszkoreit, Jones, Gomez, Kaiser, and Polosukhin}]{vaswani2017attention}
Ashish Vaswani, Noam Shazeer, Niki Parmar, Jakob Uszkoreit, Llion Jones, Aidan~N Gomez, {\L}ukasz Kaiser, and Illia Polosukhin. 2017.
\newblock Attention is all you need.
\newblock \emph{Advances in neural information processing systems}, 30.

\bibitem[{Wang et~al.(2022{\natexlab{a}})Wang, Yang, Men, Lin, Bai, Li, Ma, Zhou, Zhou, and Yang}]{ofa}
Peng Wang, An~Yang, Rui Men, Junyang Lin, Shuai Bai, Zhikang Li, Jianxin Ma, Chang Zhou, Jingren Zhou, and Hongxia Yang. 2022{\natexlab{a}}.
\newblock Ofa: Unifying architectures, tasks, and modalities through a simple sequence-to-sequence learning framework.
\newblock In \emph{International Conference on Machine Learning}, pages 23318--23340. PMLR.

\bibitem[{Wang et~al.(2022{\natexlab{b}})Wang, Bao, Dong, Bjorck, Peng, Liu, Aggarwal, Mohammed, Singhal, Som et~al.}]{beitv3}
Wenhui Wang, Hangbo Bao, Li~Dong, Johan Bjorck, Zhiliang Peng, Qiang Liu, Kriti Aggarwal, Owais~Khan Mohammed, Saksham Singhal, Subhojit Som, et~al. 2022{\natexlab{b}}.
\newblock Image as a foreign language: Beit pretraining for all vision and vision-language tasks.
\newblock \emph{arXiv preprint arXiv:2208.10442}.

\bibitem[{Wei et~al.(2022)Wei, Wang, Schuurmans, Bosma, Xia, Chi, Le, Zhou et~al.}]{wei2022chain}
Jason Wei, Xuezhi Wang, Dale Schuurmans, Maarten Bosma, Fei Xia, Ed~Chi, Quoc~V Le, Denny Zhou, et~al. 2022.
\newblock Chain-of-thought prompting elicits reasoning in large language models.
\newblock \emph{Advances in Neural Information Processing Systems}, 35:24824--24837.

\bibitem[{Yang et~al.(2022)Yang, Gan, Wang, Hu, Lu, Liu, and Wang}]{pica}
Zhengyuan Yang, Zhe Gan, Jianfeng Wang, Xiaowei Hu, Yumao Lu, Zicheng Liu, and Lijuan Wang. 2022.
\newblock An empirical study of gpt-3 for few-shot knowledge-based vqa.
\newblock In \emph{Proceedings of the AAAI Conference on Artificial Intelligence}, 3, pages 3081--3089.

\bibitem[{Yao et~al.(2023{\natexlab{a}})Yao, Yu, Zhao, Shafran, Griffiths, Cao, and Narasimhan}]{tot}
Shunyu Yao, Dian Yu, Jeffrey Zhao, Izhak Shafran, Thomas~L. Griffiths, Yuan Cao, and Karthik Narasimhan. 2023{\natexlab{a}}.
\newblock Tree of thoughts: Deliberate problem solving with large language models.
\newblock \emph{arXiv preprint arXiv:2305.10601}.

\bibitem[{Yao et~al.(2023{\natexlab{b}})Yao, Li, and Zhao}]{yao2023beyond}
Yao Yao, Zuchao Li, and Hai Zhao. 2023{\natexlab{b}}.
\newblock Beyond chain-of-thought, effective graph-of-thought reasoning in large language models.
\newblock \emph{arXiv preprint arXiv:2305.16582}.

\bibitem[{Zhang et~al.(2023)Zhang, Zhang, Li, Zhao, Karypis, and Smola}]{mmcot}
Zhuosheng Zhang, Aston Zhang, Mu~Li, Hai Zhao, George Karypis, and Alex Smola. 2023.
\newblock Multimodal chain-of-thought reasoning in language models.
\newblock \emph{arXiv preprint arXiv:2302.00923}.

\end{thebibliography}
\bibliographystyle{acl_natbib}

\newpage
\appendix

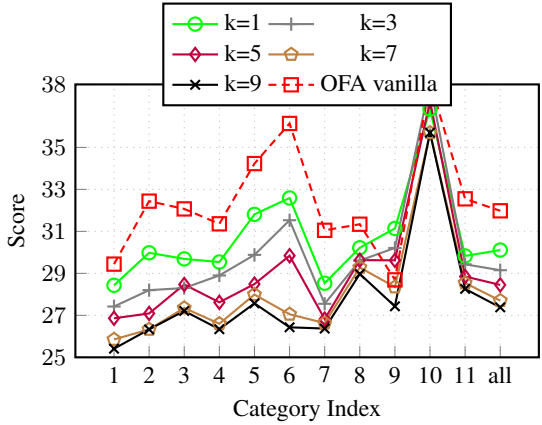
\begin{figure}[t] 
\centering 
\pgfplotsset{height=5.2cm,width=7.68cm,compat=1.14,every axis/.append style={thick},every tick label/.append style={font=\small},every axis legend/.append style={ at={(0.5,0.95)}},legend columns=2 row=2} 
    \begin{tikzpicture} \tikzset{every node}=[font=\small] 
        \begin{axis} [ 
                align = center, 
                legend style={at={(0.5,1.3)},anchor=north}, 
                ymin=25, ymax=38, 
                xticklabels={1,2,3,4,5,6,7,8,9,10,11,all}, xtick={0,1,2,3,4,5,6,7,8,9,10,11}, 
                ylabel style={align=center}, xlabel={Category Index}, ylabel={Score}, ytick={25, 27, 29, 31, 33, 35,38}, 
                ymajorgrids=true, xmajorgrids=true, grid style=dotted, xtick pos=bottom, ytick pos=left, 
            ]

        \addplot+ [color=green, mark=o, mark size=2.5pt, ] coordinates {(0, 28.43) (1, 29.98) (2, 29.69) (3, 29.54) (4, 31.81) (5, 32.59) (6, 28.53) (7, 30.22) (8, 31.14) (9, 36.82) (10, 29.83) (11, 30.11)}; \addlegendentry{k=1}

        
        \addplot+ [color=gray, mark=+, mark size=2.5pt, ] coordinates {(0, 27.42) (1, 28.2) (2, 28.3) (3, 28.9) (4, 29.88) (5, 31.53) (6, 27.55) (7, 29.61) (8, 30.21) (9, 38.04) (10, 29.43) (11, 29.15)}; \addlegendentry{k=3}
        
        
        \addplot+ [color=purple, mark=diamond, mark size=2.5pt, ] coordinates {(0, 26.86) (1, 27.09) (2, 28.47) (3, 27.62) (4, 28.49) (5, 29.84) (6, 26.82) (7, 29.63) (8, 29.62) (9, 37.21) (10, 28.85) (11, 28.45)}; \addlegendentry{k=5}
        
        \addplot+ [color=brown, mark=pentagon, mark size=2.5pt, ] coordinates {(0, 25.85) (1, 26.33) (2, 27.35) (3, 26.6) (4, 27.99) (5, 27.05) (6, 26.63) (7, 29.28) (8, 28.35) (9, 35.7) (10, 28.54) (11, 27.7)}; \addlegendentry{k=7}
        
        \addplot+ [color=black, mark=x, mark size=2.5pt, ] coordinates {(0, 25.41) (1, 26.33) (2, 27.21) (3, 26.34) (4, 27.57) (5, 26.43) (6, 26.37) (7, 28.97) (8, 27.43) (9, 35.7) (10, 28.26) (11, 27.38)}; \addlegendentry{k=9}
        
        \addplot+ [color=red, mark=square, mark size=2.5pt, ] coordinates {(0, 29.44) (1, 32.44) (2, 32.07) (3, 31.36) (4, 34.23) (5, 36.14) (6, 31.05) (7, 31.34) (8, 28.69) (9, 38.1) (10, 32.55) (11, 31.98)}; \addlegendentry{OFA vanilla}
        
        \end{axis} 
    \end{tikzpicture} \vspace{-1mm} 
\caption{The score of each category for different retrieved K using MoR on the OKVQA dataset using OFA. Category indexes are shown in Table \ref{tab:okvqa-cate}.}
\label{fig:curve-dedtails} \vspace{-2mm} 
\end{figure}

\section{Case study on NLVR2}
\label{sec:casestudy}
\begin{figure*}[!ht]
\centering
\includegraphics[width=1.0\textwidth, center]{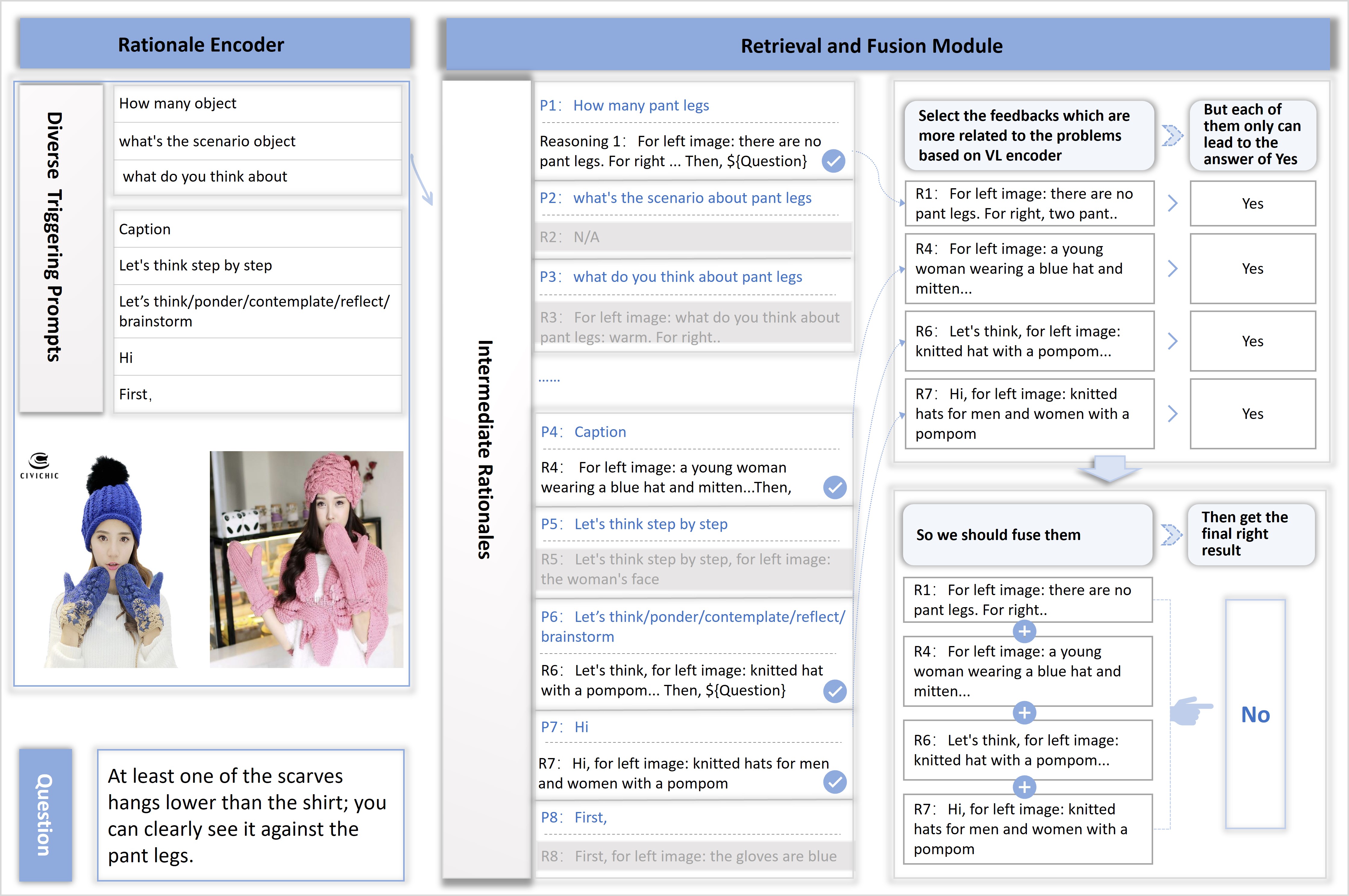}
\caption{Case Study of MoR system }
\label{fig:casestudy}
\end{figure*}

In Figure \ref{fig:casestudy}, we present a case study to demonstrate how our system works and how each module plays a role. 
\begin{itemize}
    \item The question is: "\emph{At least one of the scarves hangs lower than the shirt; you can clearly see it against the pant legs.}" 
     \item Some of the rationales $R$ are not relevant or useful so they have been filtered by the system, such as "\emph{what do you think about pant legs: warm}". However, some rationales provide a degree of relevance to the problem. For instance, "\emph{a young woman wearing a blue hat}" does not indicate if there is a scarf, but gives us some context about the scene. 
    \item These remaining relevant rationales $R$ alone may result in wrong answers, but forming multi-modal thoughts and retrieving and fusing them can get the right answer.
\end{itemize}

\section{Extended analysis for categories of OKVQA}
\label{sec:ea-okvqa}
OKVQA has ten categories as shown in table \ref{tab:okvqa-cate}. As shown in \ref{fig:curve-dedtails} We analyze the scores of each category with MoR and have these findings:
\begin{itemize}
    \item Only index 9 appears some scores surpass the red line with a square dot, which means the improvement of MoR is only observed in the category of science and technology(OKVQA-S).
    \item The other categories may be relatively easy to answer without intermediate rationales.
    \item For the category of Weather and Climate, we can see that MoR with different retrieved K and Vanilla are almost the same score, which means the answer depends more on the backbone itself to get an improvement, not the reasoning strategy.
\end{itemize}

\begin{table}[h]
\centering
\begin{tabular}{|c|p{0.8\textwidth}|}
\hline
\textbf{Index} & \textbf{Category} \\ \hline
1 & Vehicles and Transportation \\ \hline
2 & Brands, Companies and Products \\ \hline
3 & Objects, Material and Clothing \\ \hline
4 & Sports and Recreation \\ \hline
5 & Cooking and Food \\ \hline
6 & Geography, History, Language and Culture \\ \hline
7 & People and Everyday life \\ \hline
8 & Plants and Animals \\ \hline
9 & Science and Technology \\ \hline
10 & Weather and Climate \\ \hline
11 & Other \\ \hline
\end{tabular}
\caption{OKVQA categories}
\label{tab:okvqa-cate}
\end{table}

\begin{table}[t]
    \centering
    \begin{tabular}{|c|l|l|}
    \hline
    \textbf{Index} & \textbf{Triggering {Prompt}} & \textbf{Category}\\ \hline
    0 & Let's consider on & Generic \\ \hline
    1 & What the scenario about & Generic\\ \hline
    2 & Let's ponder on & Generic\\ \hline
    3 & Let's reflect on & Generic\\ \hline
    4 & Let's brainstorm on & Generic\\ \hline
    5 & What do you think on & Generic\\ \hline
    6 & Let's contemplate on & Specific\\ \hline
    7 & First, & Specific\\ \hline
    8 & Let's think & Specific\\ \hline
    9 & Hi & Specific\\ \hline
    10 & Caption & Specific\\ \hline
    \end{tabular}
    \caption{Table of Triggering Prompts: his table presents two types of triggering prompts previously explained in the \hyperref[sec:reg]{\textcolor{blue}{{rationales generation}}} section.}
    \label{tab:triggering_prompts}
    \end{table}

\section{Rationale Index}
\label{sec:rationale_index}
We have listed a portion of our triggering prompts in table \ref{tab:triggering_prompts}.

\end{document}